\documentclass[letterpaper, 10 pt, conference]{ieeeconf}  

\IEEEoverridecommandlockouts                              
\usepackage{array} 
\usepackage{amsmath} 
\usepackage{booktabs} 
\usepackage{algorithm}
\usepackage{algpseudocode}
\usepackage{graphicx}
\usepackage{subcaption}
\usepackage{amssymb}
\usepackage{comment}

\usepackage[T1]{fontenc}

\title{\LARGE \bf SPLC: Social Preference Learning for Crowd Robot Navigation}

\author{Zixuan Chen, Hao Fu$^{*}$, Haiwen Hu, Shiquan Zheng 
\thanks{This work was supported in part by the National Natural Science Foundation of China under Grant 62303357 and Grant 62173262 and in part by the Hubei Provincial Natural Science Foundation of China under Grant 2023AFB109.}
\thanks{*Corresponding author: Hao Fu}
\thanks{Z. Chen, H. Fu, H. Hu, and S. Zheng are with the School of Computer Science and Technology, Wuhan University of Science and Technology and also with the Hubei Province Key Laboratory of Intelligent Information Processing and Real-time Industrial System, Wuhan 430081, China (e-mail: fuhao@wust.edu.cn).}
}

\begin{document}

\maketitle
\thispagestyle{empty}
\pagestyle{empty}

\begin{abstract}

Offline reinforcement learning (RL) holds significant potential for crowd robot navigation in human-robot coexistence applications. However, the inherent complexity of pedestrian motion renders the design of effective reward functions for promoting socially compliant robot behaviors a persistent challenge. This paper proposes a Social Preference Learning for Crowd Robot Navigation (SPLC) algorithm to eliminate the need for detailed reward design. Its core innovation lies in the introduction of a social preference feedback mechanism to automatically generate preference data through principled preference evaluation criteria. By explicitly accounting for the intricacies of pedestrian dynamics, the pipeline mitigates the reward bias and facilitates the systematic quantification of broad social norms, thereby fostering socially compliant behaviors. Extensive experiments integrating SPLC with offline RL methods demonstrate consistent improvements over state-of-the-art baselines across standard performance metrics. Furthermore, real-world experiments on the TurtleBot4 further validate the effectiveness of SPLC in practical human–robot coexistence settings. Our code and video demos are available at https://github.com/sklus949/SPLC.
\end{abstract}

\section{INTRODUCTION}

Mobile robots are becoming increasingly prevalent in various engineering applications, such as autonomous driving and logistics distribution, which often involve pedestrian-rich environments. A key enabling technology for these applications is crowd robot navigation, aiming to avoid potential collisions and reach the target in minimal time. However, the inherent unpredictability and often uncooperative dynamics of pedestrian behavior pose serious threats to the safety of human–robot interaction.

Despite these challenges, research on crowd robot navigation has achieved substantial progress. Existing solutions can be broadly classified into three categories. One category encompasses reactive approaches, such as the Social Force Model (SFM) \cite{HelbingD1995}, Reciprocal Velocity Obstacles (RVO) \cite{van2008reciprocal}, and Optimal Reciprocal Collision Avoidance (ORCA) \cite{VanD2011}, which determine the robot’s optimal actions based on physical and geometric interaction rules. Another line of work focuses on predicting pedestrians’ future trajectories before planning the path \cite{KretzschmarH2016} \cite{TrautmanP2013}. However, both approaches are prone to the robot freezing problem in highly dynamic crowd environments.

Another approach is deep reinforcement learning (DRL) \cite{ChoiJ2021}, which holds great potential for crowd robot navigation and can overcome the limitations of the above methods. Owing to its inherent interactive nature, DRL-based crowd robot navigation has achieved substantial progress. In this process, many deep learning techniques, such as long short-term memory networks (LSTM) \cite{everett2018motion}, self-attention mechanisms \cite{chen2019crowd}, and Transformers \cite{LiuS2024}, have played important roles in encoding interactive feature information.  
However, the deployment of online DRL in real-world crowd navigation remains hindered by substantial challenges, particularly the high safety risks and considerable costs associated with human-in-the-loop training. To mitigate these limitations, offline RL has attracted increasing attention as a viable alternative. Representative algorithms, including Implicit Q-Learning (IQL) \cite{kostrikov2021offline}, Conservative Q-Learning (CQL) \cite{kumar2020conservative}, and TD3BC \cite{fujimoto2021minimalist}, have demonstrated significant potential in autonomous driving and robotics applications \cite{kiran2021deep, li2022hierarchical}. Building on these foundations, recent efforts \cite{wu2023risk,weerakoon2024vapor} have begun to explore the integration of offline RL into crowd robot navigation, highlighting a promising direction for advancing both the safety and practicality of socially compliant robot systems.

Despite significant progress about the Offline RL-based crowd robot navigation, the design of its reward function remains a primary challenge. Existing approaches predominantly rely on manually handcrafted reward functions, exhibiting an inherent limitation. Specifically, the absence of a principled quantification of broad social norms in the reward functions constrains the robot's capacity to exhibit socially compliant behaviors. For instance, in the complex crowd scenario, a handcrafted reward function may successfully guide the robot around a single individual pedestrian but fail to anticipate the collective flow of a crowd or recognize socially attentive behaviors like following a group's implicit lanes, thereby producing unnatural, myopic, or socially undesirable behaviors.

Motivated by the above challenge, this paper proposes a \textbf{S}ocial \textbf{P}reference \textbf{L}earning for \textbf{C}rowd robot navigation (SPLC). This algorithm learns a reward function by leveraging trajectory preference labeling, avoiding manual design of complex rewards in the crowd robot navigation. The learned reward model is then employed in some offline RL methods, such as IQL, CQL, and TD3BC. The main contributions of this paper can be summarized as:

\begin{itemize}
\item{The proposed SPLC integrates a social preference feedback mechanism with a preference transformer to model preference rewards, thereby ensuring that the robot exhibits socially compliant behaviors.}

\item{This paper introduces a social preference feedback mechanism to automatically generate preference labels via preference evaluation criteria. The criteria mitigate the reward bias by explicitly incorporating the inherent unpredictability and often uncooperative dynamics of pedestrian motion, thereby contributing to the quantification of broad social norms.}

\item{We evaluate our SPLC integrated with standard offline RL methods. Experimental results demonstrate that our approach significantly outperforms the baselines.}
\end{itemize}

\section{RELATED WORKS}
\subsection{Crowd Robot Navigation}
Early research on crowd robot navigation primarily focused on decoupled models, treating pedestrians as autonomous agents with independent motion patterns or as dynamic obstacles \cite{mitsch2013provably, ichter2018learning}. Some existing studies \cite{biswas2022socnavbench, xie2023drl} have identified two main limitations of these approaches: they neglect human–robot interaction modeling, which can cause robot "freezing," and they depend on human modeling, preventing autonomous adaptation to complex environments. With its increasing popularity, researchers have started shifting their focus toward learning-based approaches. Liu \textit{et al.} \cite{liu2021decentralized} proposed decentralized structural-Recurrent Neural Network (DS-RNN), which was integrated with DRL to perform spatiotemporal reasoning in crowd navigation. Mun \textit{et al.} \cite{mun2022occlusion} introduced an occlusion-aware DRL framework, integrating social occlusion inference with a variational autoencoder for safer navigation. Additionally, the integration of attention mechanisms with RL has gained traction in recent studies \cite{zhou2022navigating, vaswani2017attention, wang2023navistar} to enhance mobile robot navigation strategies.

Solving the problem of crowd robot navigation using RL requires an appropriate reward function to guide the robot's movement. Initially, the reward function used sparse rewards, rewarding task completion while penalizing collisions or uncomfortable distances, with zero rewards in other situations \cite{everett2018motion,chen2019crowd}. Due to the lack of reward signals, this approach often leads to inefficient learning. Later research introduced potential-based rewards \cite{liu2021decentralized, mun2022occlusion}, which guided the robot towards the goal by calculating the goal distance. Although this addressed the issue of sparse rewards, over-reliance on the distance may result in unnatural or myopic behaviors. More recent efforts have explored carefully engineered, scenario-specific reward functions \cite{zhou2022navigating, jiang2023learning}, achieving improved navigation performance but at the cost of extensive manual design effort. In contrast, this work advances the field by employing the proposed SPLC algorithm to model the reward for crowd-robot navigation, thereby reducing the dependence on handcrafted designs and enabling more adaptive and socially compliant behaviors.

\subsection{Preference-based reinforcement learning}
Preference-based Reinforcement Learning (PbRL) is an RL approach that learns an implicit reward function from comparative feedback on different trajectories provided by humans or experts, rather than relying on manually designed explicit reward signals \cite{christiano2017deep}. In this way, PbRL is able to capture human preferences and values, thereby guiding the agent to learn behaviors that better align with desired outcomes \cite{abdelkareem2022advances, kaufmann2024survey}. As one of the core technologies driving ChatGPT 3, this approach has attracted extensive research interest and attention \cite{du2023chat}. Several studies \cite{pacchiano2021dueling, wang2022skill, cao2024reinforcement} have successfully employed online PbRL to learn reward functions, which enable agents to act in accordance with human preferences. However, online PbRL requires continuously querying humans for preference feedback during training, which incurs a substantial human feedback cost. To overcome this limitation, offline preference-based learning has garnered significant research interest. Shin \textit{et al.} \cite{shin2023benchmarks} applied PbRL to offline RL, establishing a series of benchmarks and algorithms for offline PbRL. Kim \textit{et al.} \cite{kim2023preference} proposed a novel Transformer model that captures human preferences via non-Markovian reward-weighted modeling. Li \textit{et al.} \cite{li2025offline} proposed Scaling Preference, which allows humans to express the strength of preferences between trajectories, enabling more accurate reward learning from offline data. Nevertheless, even in offline RL, due to the complexity of real-world tasks and the inherent subjectivity of human annotators, humans often struggle to provide sufficient and accurate preference feedback. To address this issue, we introduce a social preference feedback mechanism that automatically generates preference labels based on preference evaluation criteria.

\section{METHODOLOGY}

\subsection{Problem Formulation}
The robot crowd navigation problem can be formulated as a sequential decision-making task.
In this paper, we model it as a Partially Observable Markov Decision Process (POMDP), defined as a tuple $(\mathcal{S}, \mathcal{A}, \mathcal{T}, R, \Omega, \mathcal{O}, \gamma)$, where $\mathcal{S}$ is the state space, $\mathcal{A}$ is the action space, $\mathcal{T}$ is the state transition probability, $R$ is the reward function, $\Omega$ is the observation space, $\mathcal{O}$ is the observation distribution, and $\gamma \in (0, 1]$ is the discount factor.

\begin{enumerate}
    \item \textit{State space $\mathcal{S}$:} In a crowd robot navigation environment, the state space at each time step consists of observable and unobservable states of the agent (robot and humans). The observable states comprise the agent's position $\mathbf{p} = [p_x, p_y]$,  velocity $\mathbf{v} = [v_x, v_y]$ and radius $\bar{r}$, while the unobservable states include the target position $\mathbf{p}_g = [g_x, g_y]$, preferred velocity $v_{pref} $ and heading angle $\theta $. To enhance the generality of the state representation, this paper adopts the robot-centric state representation \cite{ChenY2017}. 
    The transformed states of the robot and pedestrian are:
    \begin{equation}
    \begin{aligned}
        s^{r} &= [d_{g},v_{pref},\bar{r},{v}_{x},{v}_{y},{\theta } ] \\
        s^{i} &=[\tilde{p}_{x}^{i}, \tilde{p} _{y}^{i}, \tilde{v}_{x}^{i}, \tilde{v}_{y}^{i}, \bar{r}^{i}, d^{i}, \bar{r}+\bar{r}^{i}] \\
    \end{aligned}
    \end{equation}
    where $s^{r} $ and $s^{i} $ represent the transformed states of the robot and the $i$-th pedestrian, respectively. $d_{g}=\left \|\mathbf{p}_g-\mathbf{p} \right \|$ is the distance from the robot to the target position, and $d^{i}=\left \|\mathbf{p}-\mathbf{p}_i \right \|$ is the distance from the robot  to pedestrian $i$. Then, the joint state can be represented as $s_{t}^{jn} =[s_{t}^{r}, s_{t}^{1},s_{t}^{2}, \ldots]$ at the \textit{t}-th time step.

    \item \textit{Action space $\mathcal{A}$:} In social scenarios, designing an appropriate action space helps robots generate smooth trajectories. Considering the kinematic specifications of real robots and their practical applications, this paper defines the robot’s actions as:
    \begin{equation}
    a_{t}=(v_{t}, \omega_{t})
    \end{equation}
    where $v_t \in (0,1) $ is the translational velocity  and $\omega_t \in (-1,1)$ is the rotational velocity.

    \item \textit{Optimization objective:} The objective of crowd robot navigation is to reach the target in the shortest possible time while avoiding potential collisions, thereby accomplishing the navigation task in an efficient and safe manner. Within the MDP framework, the objective is generally formulated via reward modeling, with the optimal policy expressed as:
    \begin{equation}
         \pi^{*} = \arg \max_{\pi}  \sum_{t} \gamma^{t} r_{t}
    \end{equation}
    where $\gamma$ is a discount factor, $\pi : \mathcal{S} \to P(\mathcal{A})$ denotes the policy, and $r_{t}$ is the reward at timestep $t$. 
\end{enumerate}

\subsection{Social Preference Learning}
\begin{figure}[htbp]
    \centering
    \includegraphics[width=\linewidth]{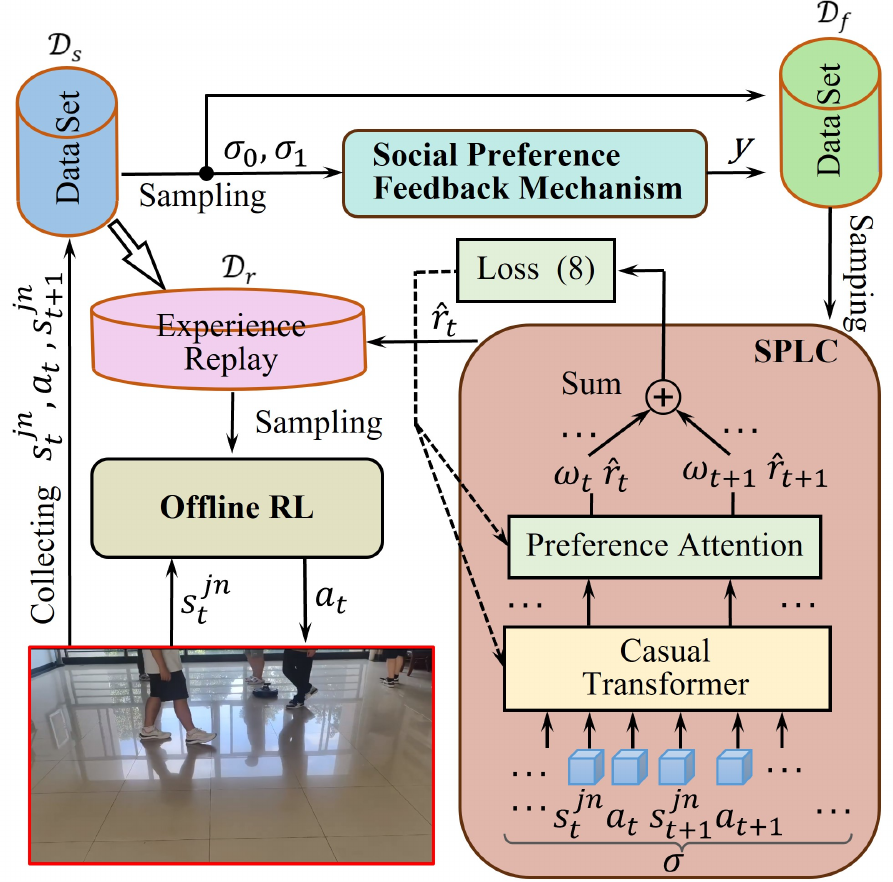}  
    \caption{The overall framework of our SPLC algorithm }
    \label{framework}
\end{figure}

The central challenge in achieving effective crowd robot navigation lies in the elaborate design of a reward function capable of guiding the learning process toward an optimal policy. In fact, handcrafted reward functions often struggle to quantify broad social norms.  As a result, RL-based on such manually specified rewards frequently leads to suboptimal navigation policies, ultimately limiting the robot’s ability to exhibit socially compliant behaviors. To address this issue, this paper proposes the SPLC algorithm to circumvent manual reward engineering by leveraging trajectory preference labeling. Its concrete overall framework is illustrated in Fig.~\ref{framework}.

PbRL typically learns a reward function from human-labeled preference data prior to the RL training phase. In this paradigm, a dataset annotated with human preference feedback is first collected and then used to guide the subsequent learning process. Nevertheless, this pipeline is associated with substantial annotation costs, as it requires extensive human labor. On the other hand, the inherent unpredictability and often uncooperative dynamics of pedestrian motion aggravate the reward bias, resulting from the inherent subjectivity of human annotators. 

To overcome these limitations, this paper introduces a social preference feedback mechanism to automatically generate preference labels for each sampling instance, consisting of a pair of trajectory segments.  The trajectory segment $\sigma=\{(s_{1}^{jn}, a_{1}), \ldots, (s_{L}^{jn}, a_{L})\} \in $ is sampled from the offline dataset $\mathcal{D}_{s}=\{s_{t}^{jn}, a_{t}, s_{t+1}^{jn} \}_{t=1}^{M_s}$ with the maximum capacity $M_s$, is a time-indexed sequence of joint states and actions within a specified length $L$. A pair of randomly sampled trajectory segments is represented as $\sigma^{0}$ and $\sigma^{1}$, respectively. The associated preference label $y$ takes the form of a two-dimensional tuple, selected from the set $y \in \{(0.9, 0.1), (0.1, 0.9), (0.7-\alpha, 0.3+\alpha), (0.3-\alpha, 0.7+\alpha), (0.5-\alpha, 0.5+\alpha)\}$ with the preference risk $\alpha$.  To assign appropriate preference labels to $\sigma^{0}$ and $\sigma^{1}$, we define three preference evaluation criteria to determine preference labels by means of quantization of broad social norms. The criteria provide structured supervisory signals, and their integration with trajectory-level preference comparisons enables the implicit inference and capture of more subtle social behaviors.

\textit{Collision Occurrence}: Indicates whether a collision between the robot and a pedestrian occurs within the trajectory segment. This serves as a direct measure of human–robot safety.

\textit{Goal Progress}: Quantifies the robot’s progress toward its goal, reflecting navigation efficiency. It is defined as
\begin{align} \label{goalpro}
\eta=\left \| \mathbf{p}_1-\mathbf{p}_g \right \|- \left \|  \mathbf{p}_L-\mathbf{p}_g\right \|,
\end{align}
where $\mathbf{p}_1$ and $\mathbf{p}_{L}$ denote the start and end robot positions of the trajectory segment $\sigma$, respectively.

\textit{Risk Exposure}: Serves as a supplement to \textit{Goal Progress} and measures the frequency of risk events, where a risk is defined as the robot entering unsafe proximity to a pedestrian. This criterion is given by
\begin{align} \label{Risk}
\mu=\frac{1}{L}\Sigma_{t=1}^{L} \xi_{t},
\end{align}
where $\xi_{t}$ is a binary indicator of the danger zone \cite{FuH2024} of a pedestrian at time step $t$. 

Owing to the fundamental importance of these segment evaluation criteria in the crowd robot navigation, we impose a lexicographic priority order on the mechanism used to determine the preference label $y$ from the 2-D tuple. The ordinal hierarchy is specified as:  "\textit{Collision Occurrence}"$\leftarrow$"\textit{Goal Progress}". 

According to this ordering, the evaluation of \textit{Collision Occurrence} has the highest priority. A segment $\sigma^{0}$ without a collision is always preferred over another segment $\sigma^{1}$ with one, i.e., $\sigma^{0} \succ \sigma^{1}$, where $\succ$ indicates a preference operator. In this case, the preference label is set to $y=(0.9, 0.1)$. Conversely, a segment $\sigma^{1}$ without a collision is always preferred over another segment $\sigma^{0}$ with one, i.e., $\sigma^{0} \prec \sigma^{1}$. Accordingly, the preference label becomes $y=(0.1, 0.9)$. If both segments about \textit{Collision Occurrence} are identical, this mechanism proceeds to the next criterion, i.e., \textit{Goal Progress}. 

The criterion of \textit{Goal Progress} is determined by the values of $\eta^{0}$ and $\eta^{1}$ as defined in Equation (\ref{goalpro}). If $\eta^{0} > \eta^{1}$, $\sigma^{0}$ is more preferable to $\sigma^{1}$, i.e., $\sigma^{0} \succ \sigma^{1}$. Afterward, we can obtain the preference label $y=(0.7-\alpha, 0.3+\alpha)$. Conversely, if $\eta^{0} < \eta^{1}$, the preference is reversed, resulting in  $\sigma^{0} \prec \sigma^{1}$ and  $y=(0.3-\alpha, 0.7+\alpha)$. If both segments are identical with respect to \textit{Goal Progress}, indicating no differential preference between the segments, a neutral label is assigned as $y=(0.5-\alpha, 0.5+\alpha)$.

\textit{Risk Exposure}, as a supplement to \textit{Goal Progress}, calculates $\mu^{0}$ and $\mu^{1}$ for each segment according to (\ref{Risk}). Then, we can have the preference risk
\begin{align} \label{PRisk}
    \alpha = 0.1 \cdot \tanh(\frac{\mu^{0}-\mu^{1}}{\mu_{\max}-\mu_{\min}}),
\end{align}
where $\mu_{\max}$ and $\mu_{\min}$ denote the maximum and minimum numbers of risk events within the segments in the dataset, respectively.

To obviate the need for designing a handcrafted rewardneed, the trajectory segments and the preference labels are used to train a reward model, serving as the downstream offline RL robot navigation task. As a result, we store the segments $(\sigma^{0}, \sigma^{1})$, and the preference label $y$ in the preference dataset $\mathcal{D}_{f} = \{\sigma_{i}^{0}, \sigma_{i}^{1}, y_{i}\}_{i=1}^{M_f}$, where $M_f$ represents the size of the dataset. 

Intuitively, socially compliant decisions inherently rely on its temporal information. Consequently, its temporal information has a role to play in designing reward functions. To this end, this paper employs the preference transformer to learn the reward function from the preference dataset  $\mathcal{D}_{f}$. The trajectory segment $\sigma$ is fed into the preference transformer to generate the weighted sum of non-Markovian rewards $\sum_{\sigma} w_{\psi,t} \hat{r}_{\psi,t}$ and the preference reward prediction sequences $[\hat{r}_{\psi,1},\hat{r}_{\psi,2},\ldots,\hat{r}_{\psi,L}]$, where $\psi$ is the trained parameter and $w_{\psi,t}$ is the importance weight. 
According to the Bradley-Terry model~\cite{RalphA1952}, we can model a preference reward predictor as follows:
\begin{equation} \label{Ppsi}
P_{\psi }(\sigma^{1} \succ \sigma^{0} ) = \frac{\mathrm{exp}(\textstyle \sum_{\sigma^{1}} w_{\psi,t} \hat{r}_{\psi,t})} {\mathrm{exp}(\textstyle \sum_{\sigma^{1}} w_{\psi,t} \hat{r}_{\psi,t})+\mathrm{exp}(\textstyle \sum_{\sigma^{0}} w_{\psi,t} \hat{r}_{\psi,t})}.
\end{equation}

Subsequently, through the preference dataset $\mathcal{D}_{f}$, the reward function $\hat{r}_{\psi,t}$ is updated by minimizing binary cross-entropy loss:
\begin{align} \label{loss}
 \mathcal{L}(\psi )=-\mathbb{E}_{( \sigma^0, \sigma^1,y)\in \mathcal{D}_{f}} & (y(0)\text{log}P_\psi(\sigma^{0}\succ\sigma^{1}) \nonumber \\ 
 & +y(1)\text{log}P_\psi(\sigma^0 \prec \sigma^1)).   
\end{align}

\textit{Remark 1:} Our proposed SPLC algorithm models the preference reward through a novel integration of a social preference feedback mechanism and the preference transformer. The social preference feedback mechanism is able to automatically generate preference labels without manual labeling by annotators through the defined preference evaluation criteria. In addition, the preference evaluation criteria can eliminate the impact on the inherent unpredictability and often uncooperative dynamics of pedestrian motion, thereby mitigating the reward bias. Consequently, our SPLC circumvents the challenges of manual reward engineering, enabling offline RL agents to acquire socially compliant behaviors.

Combining with the offline dataset $\mathcal{D}_{s}$ and the reward prediction $\hat{r}_{\psi,t}$ yields transitions $(s_{t}^{jn},a_{t},s_{t+1}^{jn},\hat{r}_{\psi,t})$. These transitions are collected into an experience replay buffer $\mathcal{D}_{r}$ to optimize robot navigation policies via offline RL algorithms,  such as IQL, CQL, and TD3BC.

The detailed implementation procedure for data collection is shown in Algorithm \ref{alg2}.

\begin{algorithm}[t]
    \caption{SPLC}
    \label{alg2}
    \begin{algorithmic}[1]
        \State \textbf{Input} Offline dataset $\mathcal{D}_{s}$, preference dataset size $M_{f}$, epochs $N_{p}$ and $N_{e}$
        \State \textbf{Initialize} Preference dataset $\mathcal{D}_{f}$, replay buffer $\mathcal{D}_{r}$, parameter $\psi$
        \While{$|\mathcal{D}_{f}|\le M_{f}$}
        \Comment{Social Preference Feedback Mechanism}
            \State Sample a pair of segments $(\sigma^{0}, \sigma^{1})\sim \mathcal{D}_{s}$
            \If{\textit{Collision occurrence}$(\sigma^{0},\sigma^{1})=$(\textit{False}, \textit{True})}
                \State  $\sigma^{0} \succ \sigma^{1}$ and $y=(0.9,0.1)$
            \ElsIf{\textit{Collision occurrence}$(\sigma^{0},\sigma^{1})=$(\textit{True}, \textit{False})}
                \State  $\sigma^{0} \prec \sigma^{1}$ and $y=(0.1,0.9)$
            \Else 
                \State Calculate $\eta^{0}$ and $\eta^{1}$ via (\ref{goalpro})
                \State Compute $\alpha$ in (\ref{Risk}) and (\ref{PRisk})
                \If{$\eta^{0}>\eta^{1}$}
                    \State  $\sigma^{0} \succ \sigma^{1}$ and $y=(0.7-\alpha,0.3+\alpha)$
                \ElsIf{$\eta^{0}<\eta^{1}$}
                    \State  $\sigma^{0} \prec \sigma^{1}$ and $y=(0.3-\alpha,0.7+\alpha)$
                \Else
                    \State $y=(0.5-\alpha,0.5+\alpha)$
                \EndIf
            \EndIf
            \State Append ($\sigma^{0}, \sigma^{1}, y$) to $\mathcal{D}_{f}$
        \EndWhile
        \For{$Step = 1\ to \ N_{p}$} \Comment{Learn Reward Model}
            \State Sample a random minibatch $(\sigma^{0}, \sigma^{1}, y)\sim \mathcal{D}_{f}$
            \State Calculate $P_{\psi }(\sigma^{1} \succ \sigma^{0} )$ in (\ref{Ppsi})
            \State Optimize $\hat{r}_{\psi,t}$ with respect to $\psi$ via Loss (\ref{loss})
        \EndFor
        \For{$Step = 1\ to \ N_{e}$} \Comment{Training Policy}
            \State Sample a random mini-batch transition tuples $(s_{t}^{jn}, a_{t}, s_{t+1}^{jn}) \sim \mathcal{D}_{s}$
            \State Compute reward $\hat{r}_{\psi,t}$ via preference transformer
            \State Append ($s_{t}^{jn}, a_{t}, s_{t+1}^{jn},\hat{r}_{\psi,t}$) to $\mathcal{D}_{r}$
            \State Update navigation policy with offline RL algorithm (e.g., IQL, CQL, TD3BC)
        \EndFor
    \end{algorithmic} 
\end{algorithm}

\section{EXPERIMENTS}
\subsection{Experiment Setup}
\begin{enumerate}

 \item \textit{Simulation environment:} This paper builds a crowd robot navigation simulation platform where a robot can navigate across diverse crowd scenarios. In the experimental setup, six pedestrians are randomly initialized within a circular area of 4 meters in radius. Each pedestrian’s initial and target positions are symmetrically distributed along the circumference of the circle. 
 Their motion behavior is governed by ORCA \cite{VanD2011}, ensuring collision-free navigation and obstacle avoidance. Unlike fixed-goal settings, pedestrians in this environment are assigned new random destinations immediately upon reaching their current targets, thereby maintaining a dynamic and continuously evolving interaction space.
 \item \textit{Generating Datasets:} Our datasets are built within this simulation environment. To evaluate the performance of the reward model in complex environments, the dataset used in this paper is of medium level. Details of the dataset, five metrics, i.e., "\textit{Success}", "\textit{Collision}", "\textit{Timeout}", "\textit{Time}" and "\textit{Capacity}" are shown in Table \ref{table1}. They describe the success rate, collision rate, timeout rate, average navigation time of success trajectories and datasets capacity, respectively.
 \begin{table}[h]
    \centering
    \caption{Specific performance metrics of datasets}
    \label{table1}
    \begin{tabular}{lccccc}
    \toprule
    \textbf{Success} & \textbf{Collision} & \textbf{Timeout} & \textbf{Time} & \textbf{Capacity} \\
    \midrule
     76.2\% & 23.6\% & 0.2\% & 11.7 s & $5 \times 10^5$ \\
    \bottomrule
    \end{tabular}
\end{table}
 \item \textit{Baseline:} To rigorously evaluate the effectiveness of our proposed algorithm, we adopt three representative offline RL methods, including IQL, CQL and TD3BC. Furthermore, to provide a comprehensive comparison of our reward function, we benchmark it against three state-of-the-art alternatives: a traditional handcrafted reward function (HR) \cite{mun2022occlusion}, a reward function derived from human-labeled preference data (HPR) \cite{wang2023navistar}, and a robust preference-based reward under corrupted preference modeling (RPR) \cite{heo2025mixing}.
 \item \textit{Training Settings:} The parameters related to preference queries and  reward model are summarized in Table \ref{table2}. In the offline RL section, we adopt the implementations of these algorithms from the CORL \cite{tarasov2023corl} library.
    
    \begin{table}[h]
    \centering
    \caption{Hyperparameters for training reward models}
    \label{table2}
    \begin{tabular}{lll}
        \toprule
        \textbf{Hyperparameter} & \textbf{Value} \\
        \midrule
         Max trajectory segments length & 100 \\
         Number of queries & 2000 \\
         Query length & 15 \\
         Number of layers & 1 \\
         Number of attention heads & 4 \\
         Embedding dimension & 256 \\
         Batch size & 256 \\
         Dropout rate & 0.1 \\
         Learning rate & $1\mathrm{e}{-4}$ \\
         Optimizer & AdamW \\
         Weight decay & $1\mathrm{e}{-4}$ \\
         Total gradient steps & $1\mathrm{e}{4}$ \\
        \bottomrule
    \end{tabular}
    \end{table}

\end{enumerate}

\begin{figure*}[t]
    \centering
    \begin{subfigure}[b]{0.23\textwidth}
        \centering
        \includegraphics[width=\textwidth]{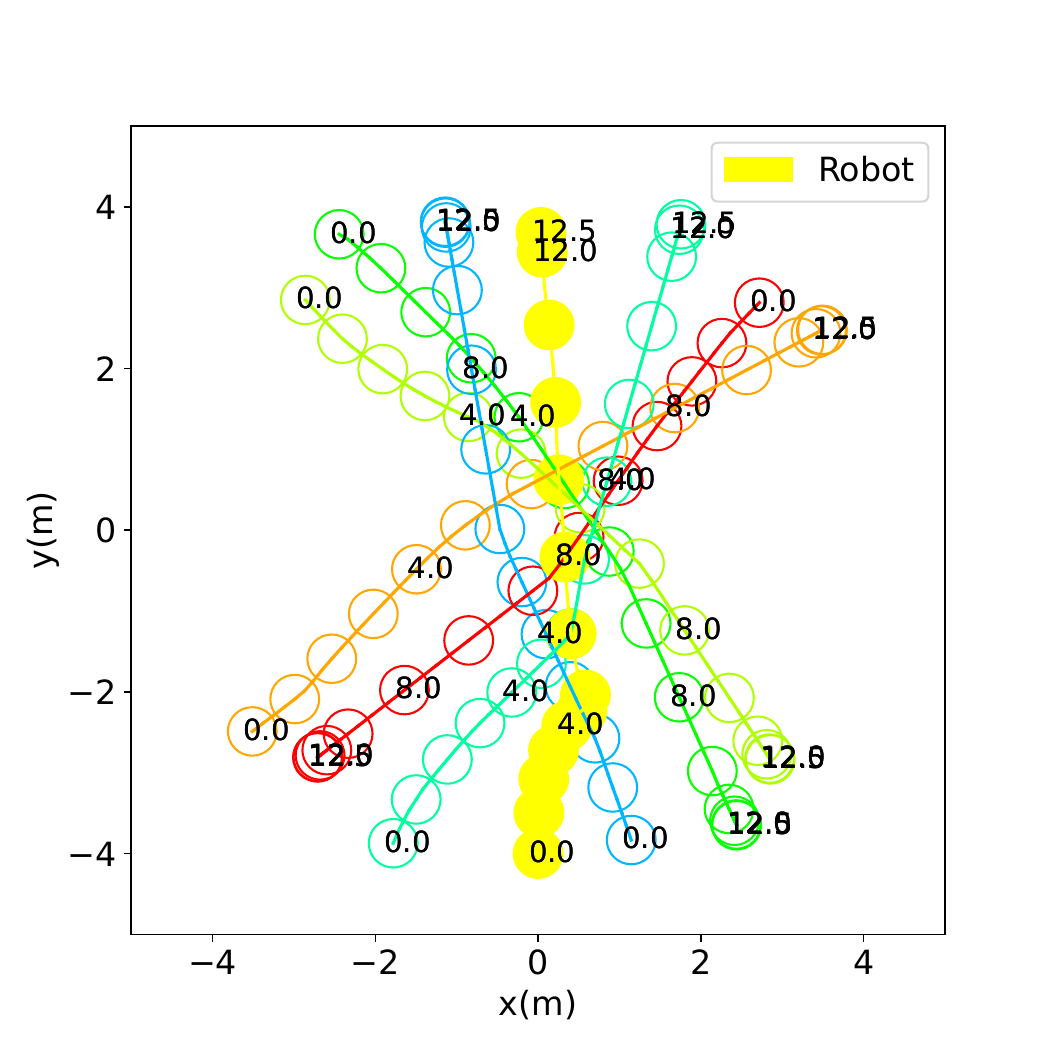}
        \caption{HR-IQL}
        \label{IQL}
    \end{subfigure}
    \hfill
    \begin{subfigure}[b]{0.23\textwidth}
        \centering
        \includegraphics[width=\textwidth]{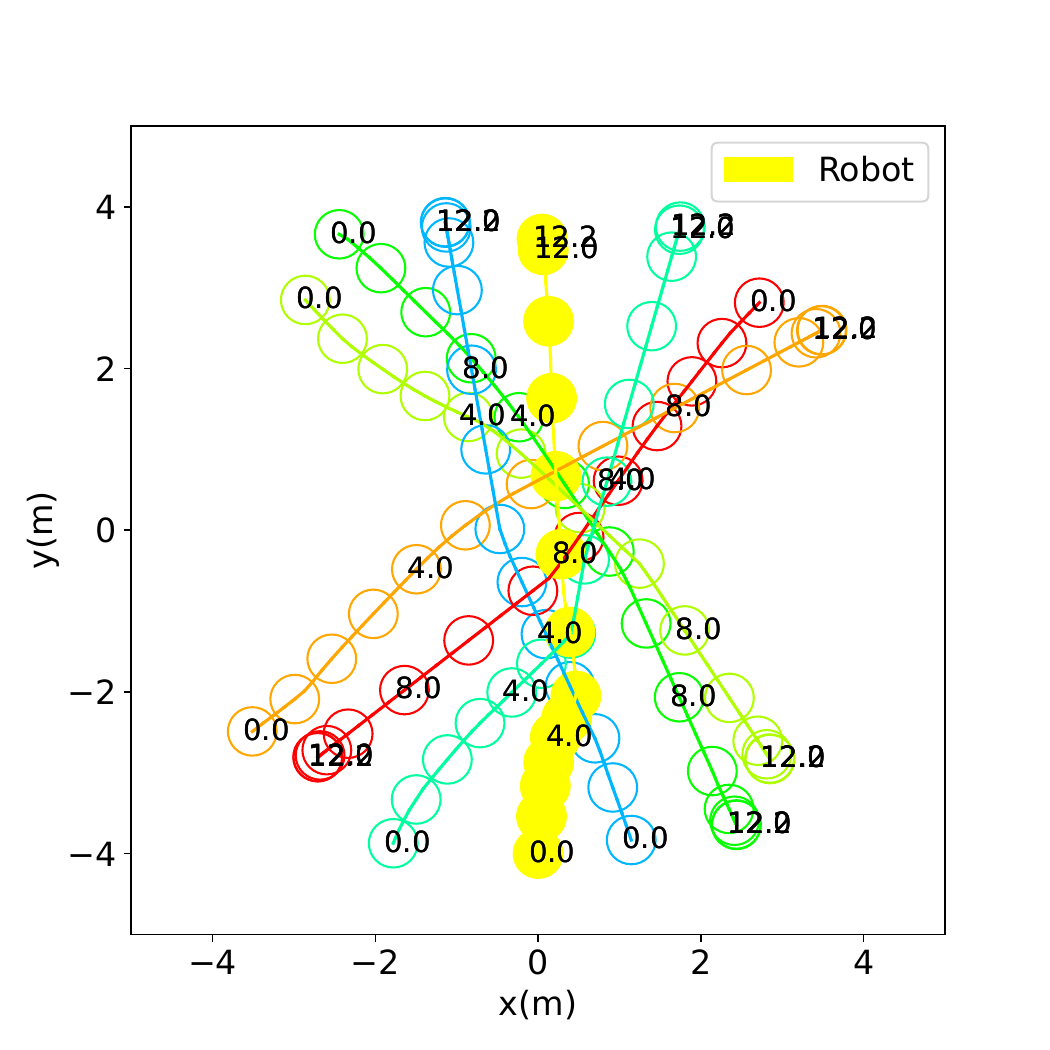}
        \caption{HPR-IQL}
        \label{HPR-IQL}
    \end{subfigure}
    \hfill
    \begin{subfigure}[b]{0.23\textwidth}
        \centering
        \includegraphics[width=\textwidth]{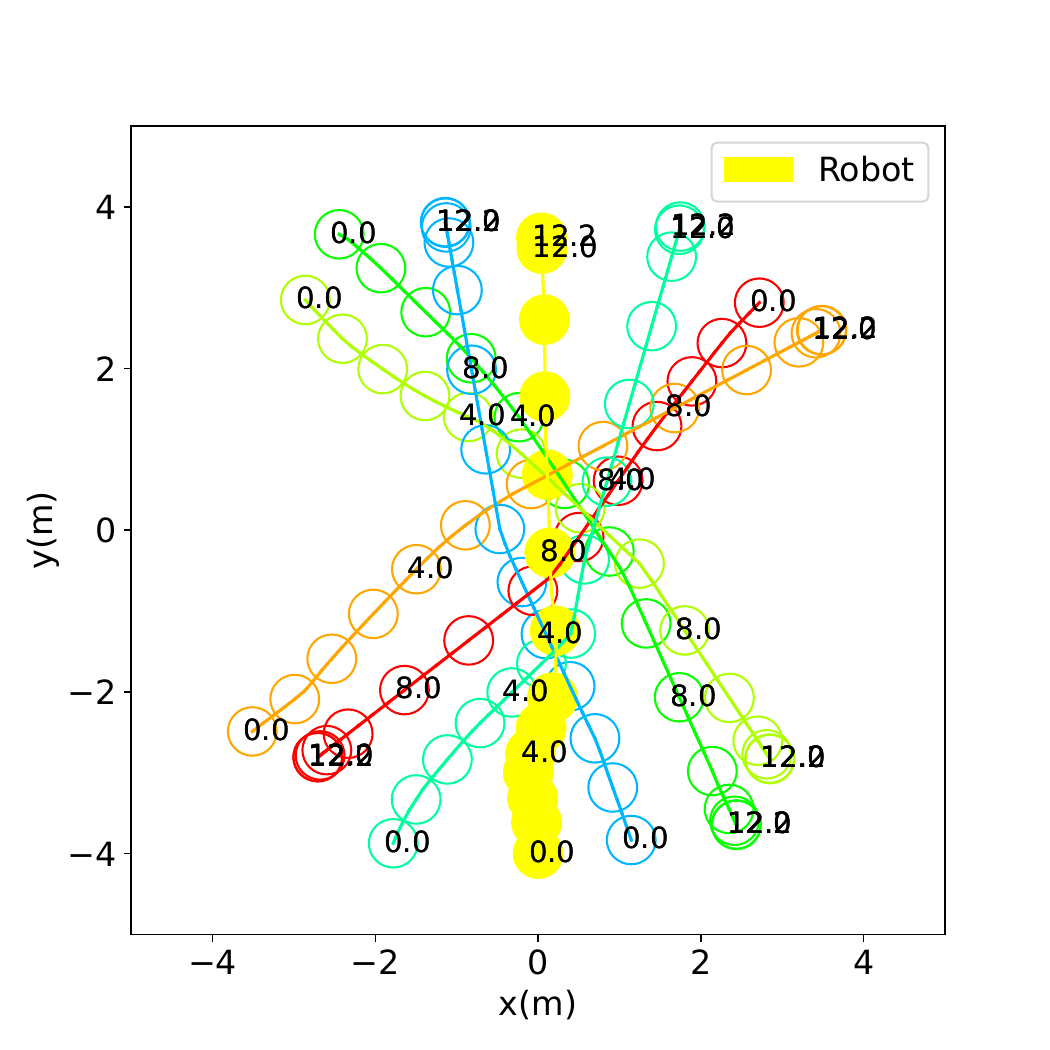}
        \caption{RPR-IQL}
        \label{RPR-IQL}
    \end{subfigure}
    \hfill
    \begin{subfigure}[b]{0.23\textwidth}
        \centering
        \includegraphics[width=\textwidth]{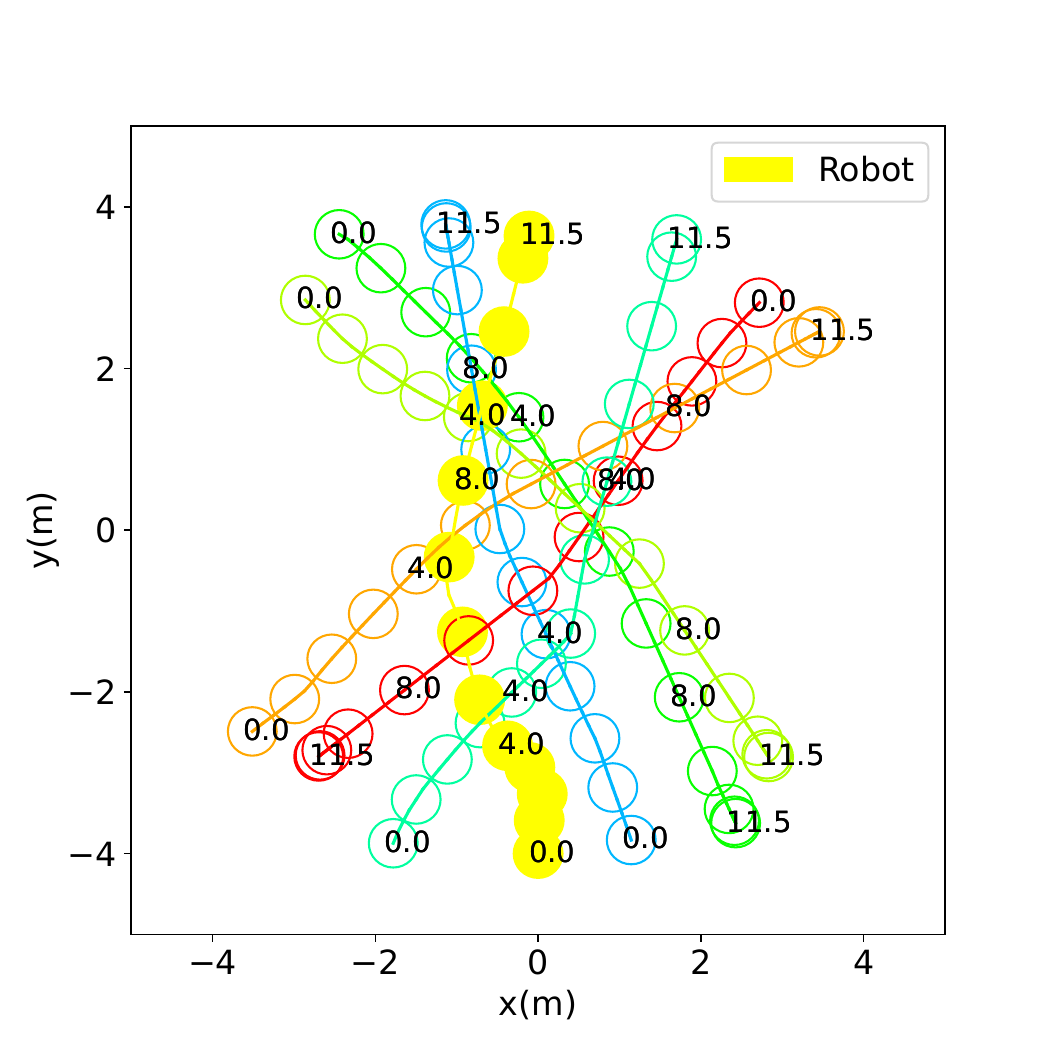}
        \caption{SPLC-IQL}
        \label{SPLC-IQL}
    \end{subfigure}


    \begin{subfigure}[b]{0.23\textwidth}
        \centering
        \includegraphics[width=\textwidth]{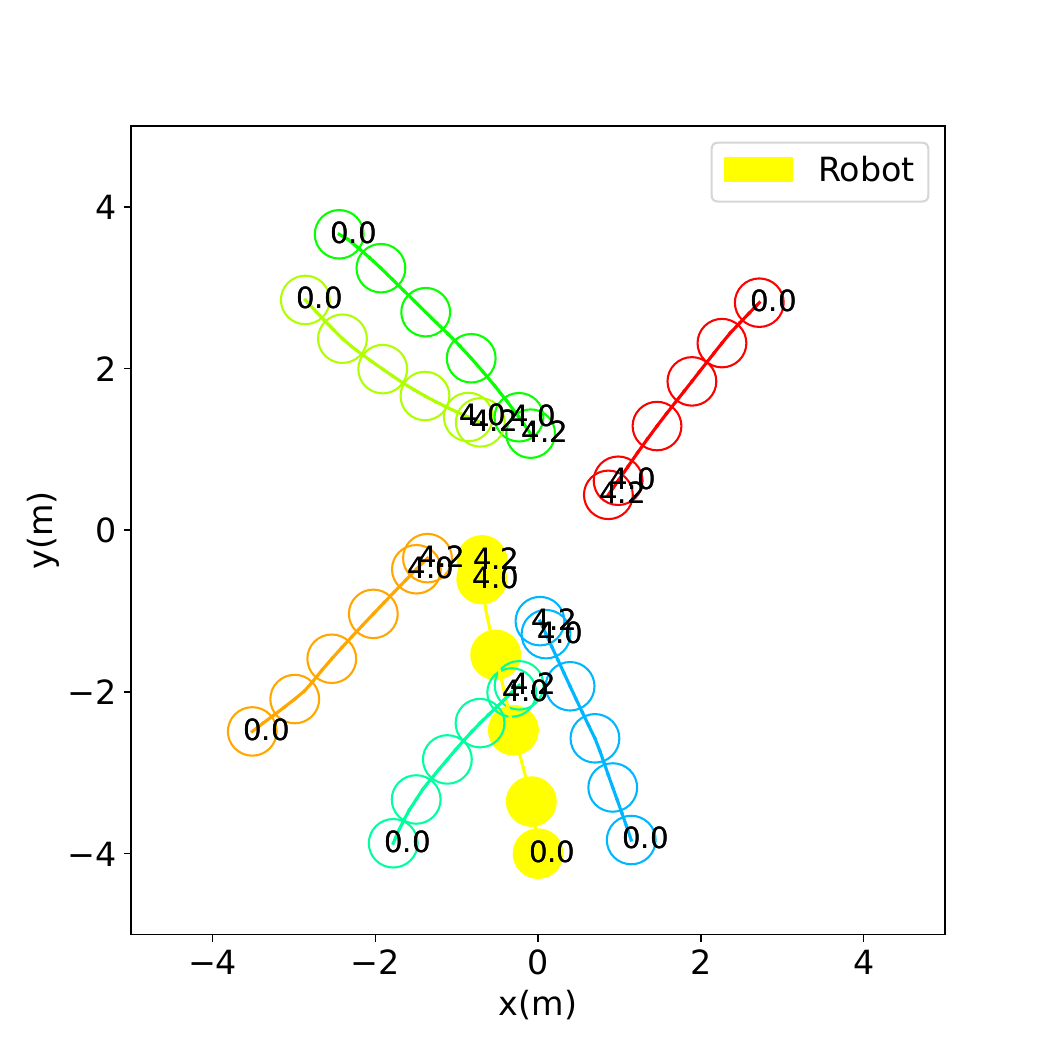}
        \caption{HR-CQL}
        \label{CQL}
    \end{subfigure}
    \hfill
    \begin{subfigure}[b]{0.23\textwidth}
        \centering
        \includegraphics[width=\textwidth]{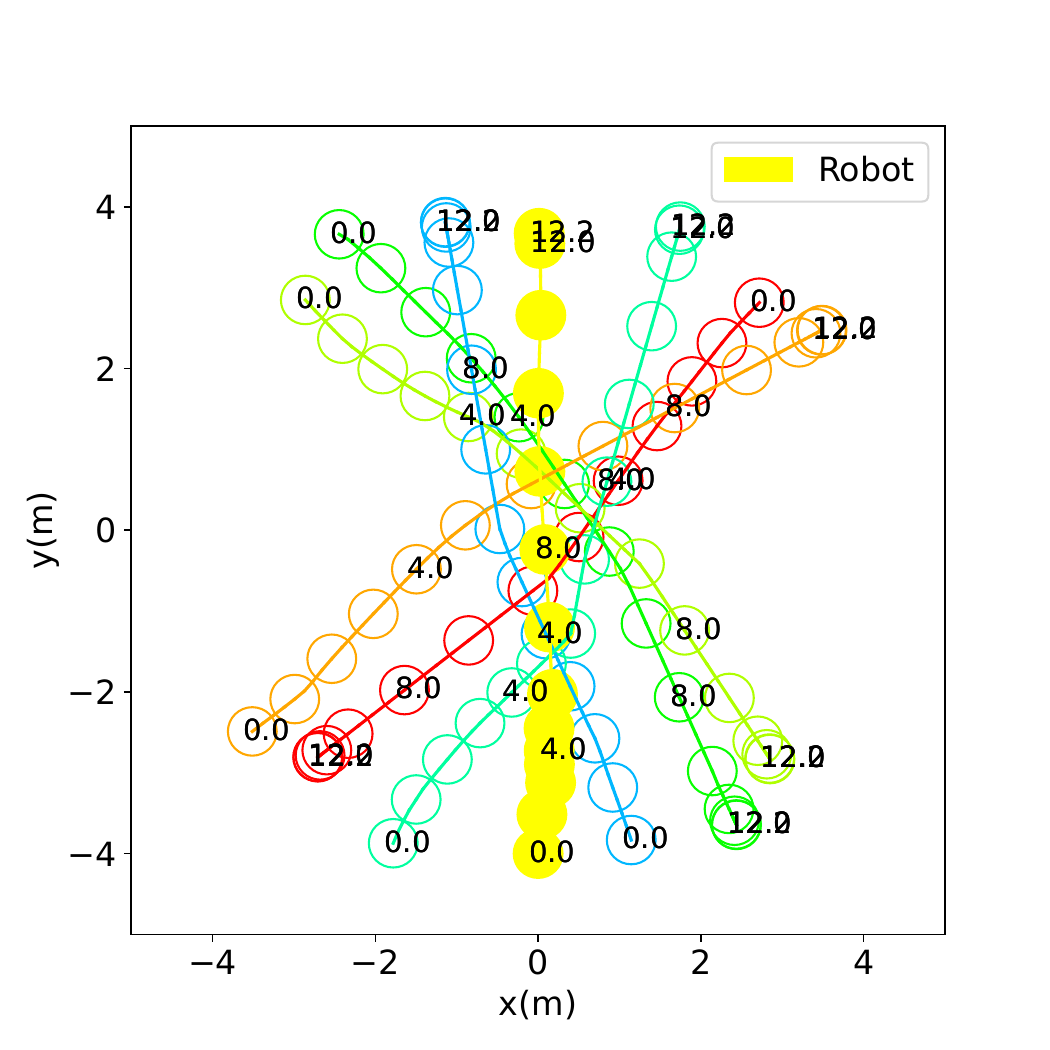}
        \caption{HPR-CQL}
        \label{HPR-CQL}
    \end{subfigure}
    \hfill
    \begin{subfigure}[b]{0.23\textwidth}
        \centering
        \includegraphics[width=\textwidth]{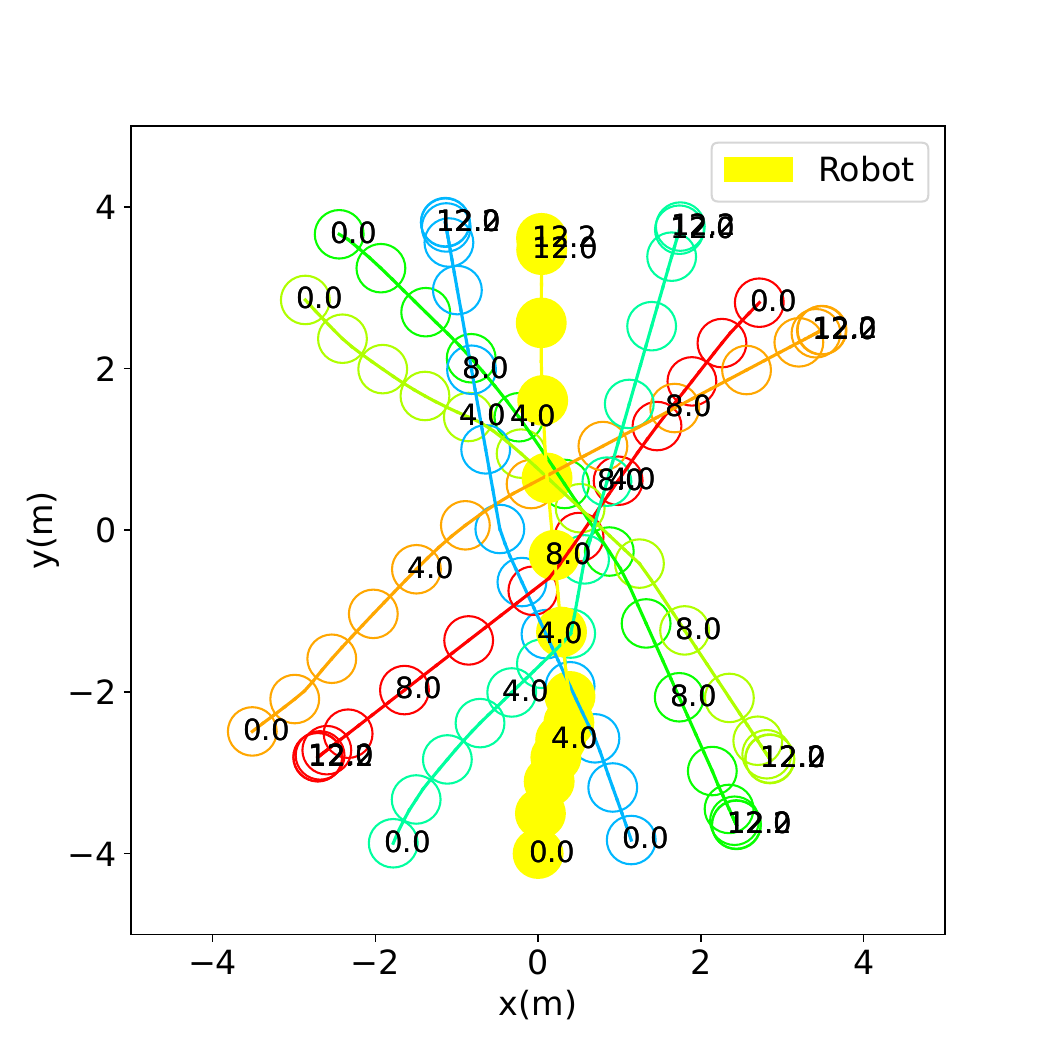}
        \caption{RPR-CQL}
        \label{RPR-CQL}
    \end{subfigure}
    \hfill
    \begin{subfigure}[b]{0.23\textwidth}
        \centering
        \includegraphics[width=\textwidth]{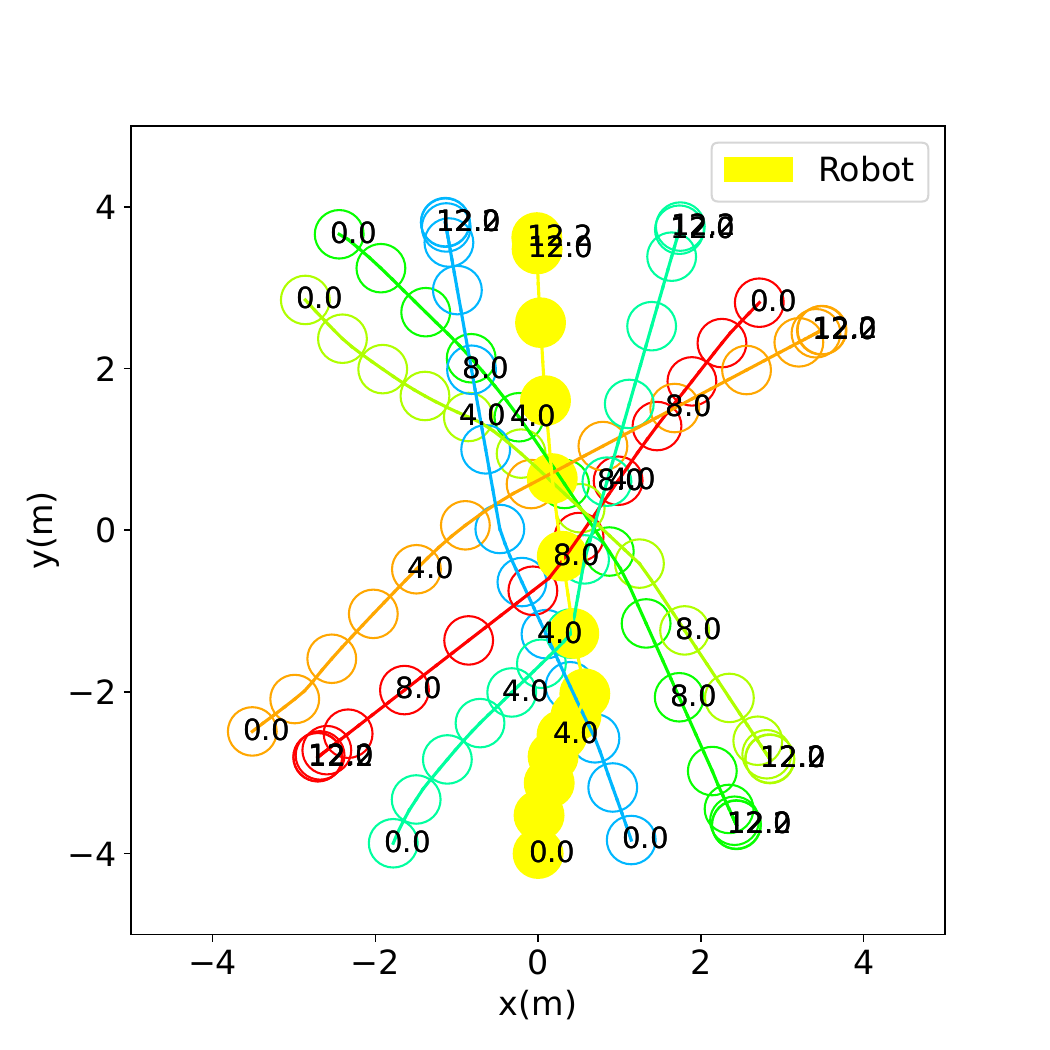}
        \caption{SPLC-CQL}
        \label{SPLC-CQL}
    \end{subfigure}
    

    \begin{subfigure}[b]{0.23\textwidth}
        \centering
        \includegraphics[width=\textwidth]{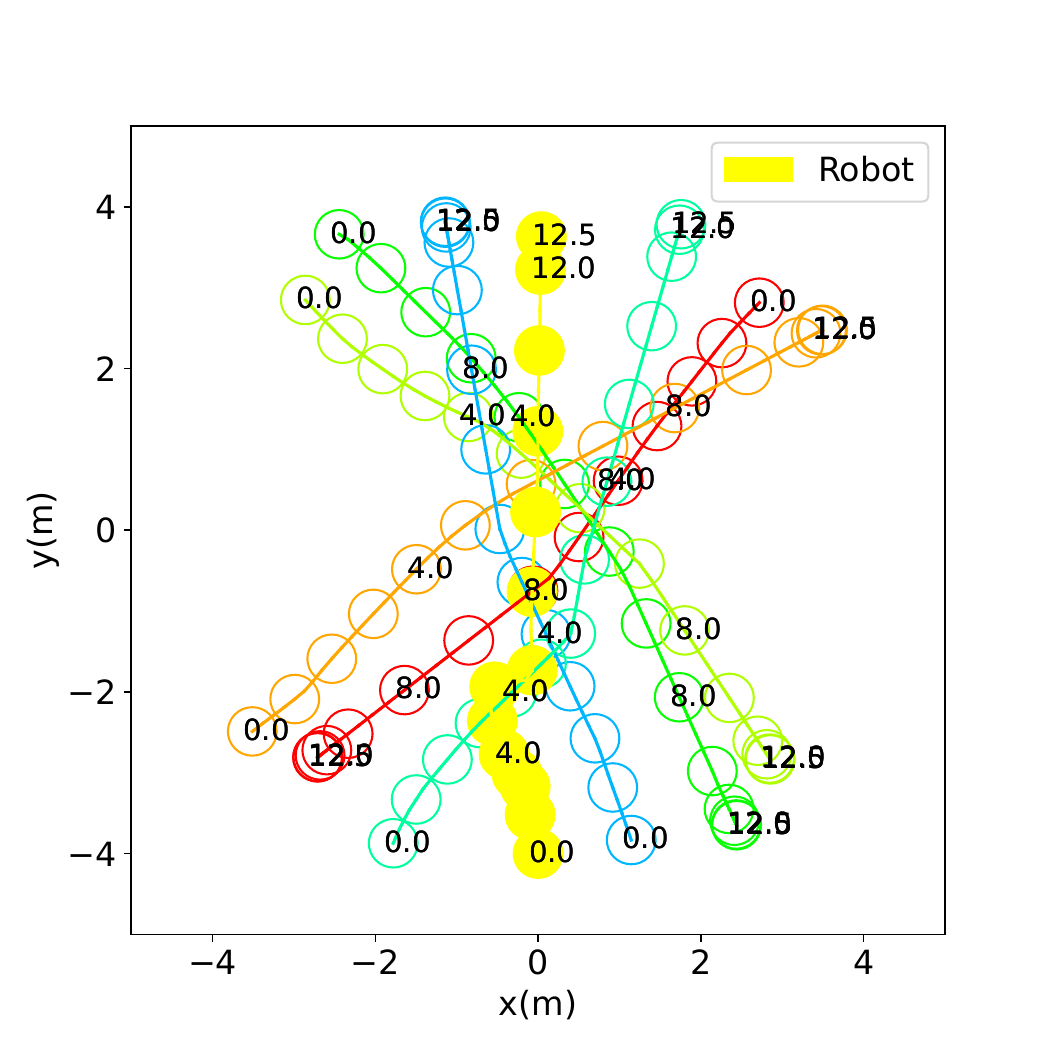}
        \caption{HR-TD3BC}
        \label{TD3BC}
    \end{subfigure}
    \hfill
    \begin{subfigure}[b]{0.23\textwidth}
        \centering
        \captionsetup{font=small}
        \includegraphics[width=\textwidth]{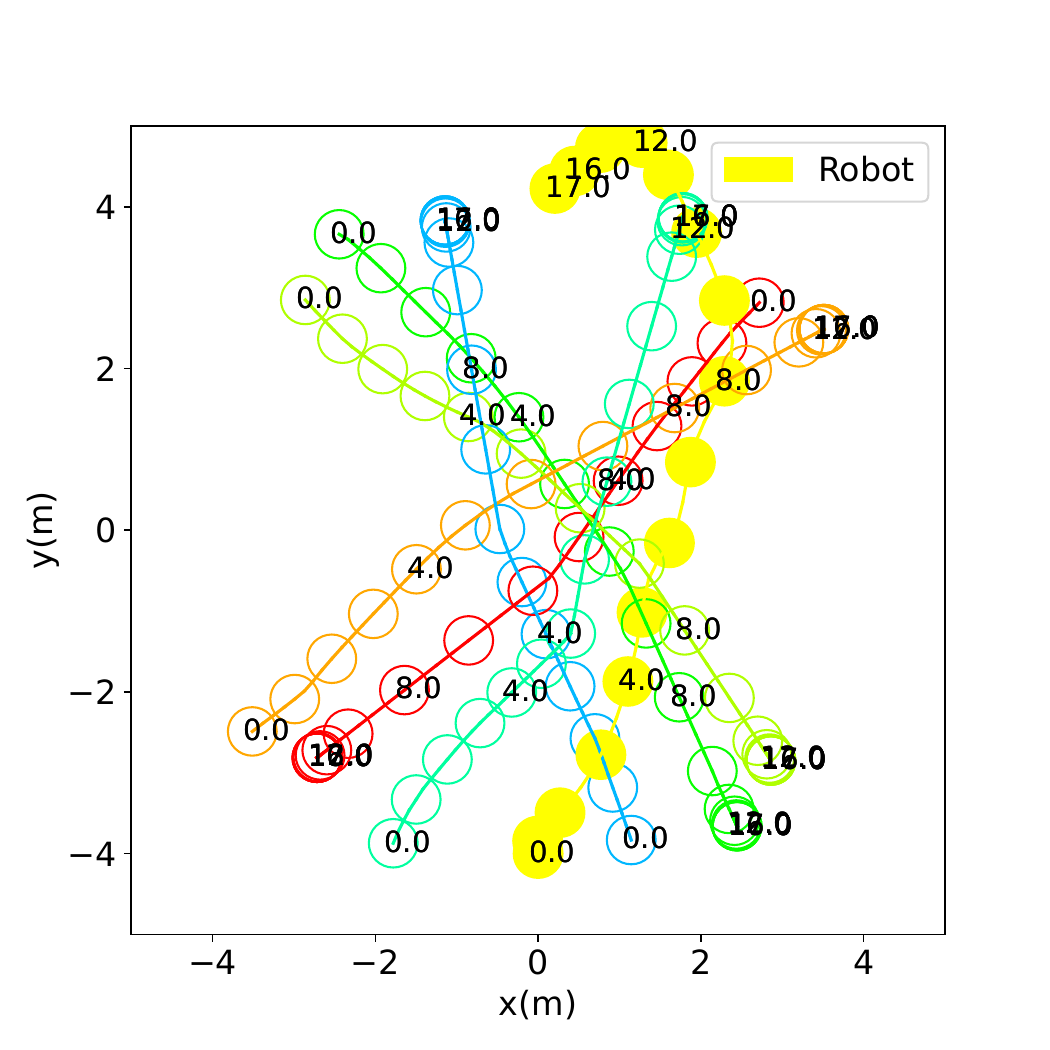}
        \caption{HPR-TD3BC}
        \label{HPR-TD3BC}
    \end{subfigure}
    \hfill
    \begin{subfigure}[b]{0.23\textwidth}
        \centering
        \includegraphics[width=\textwidth]{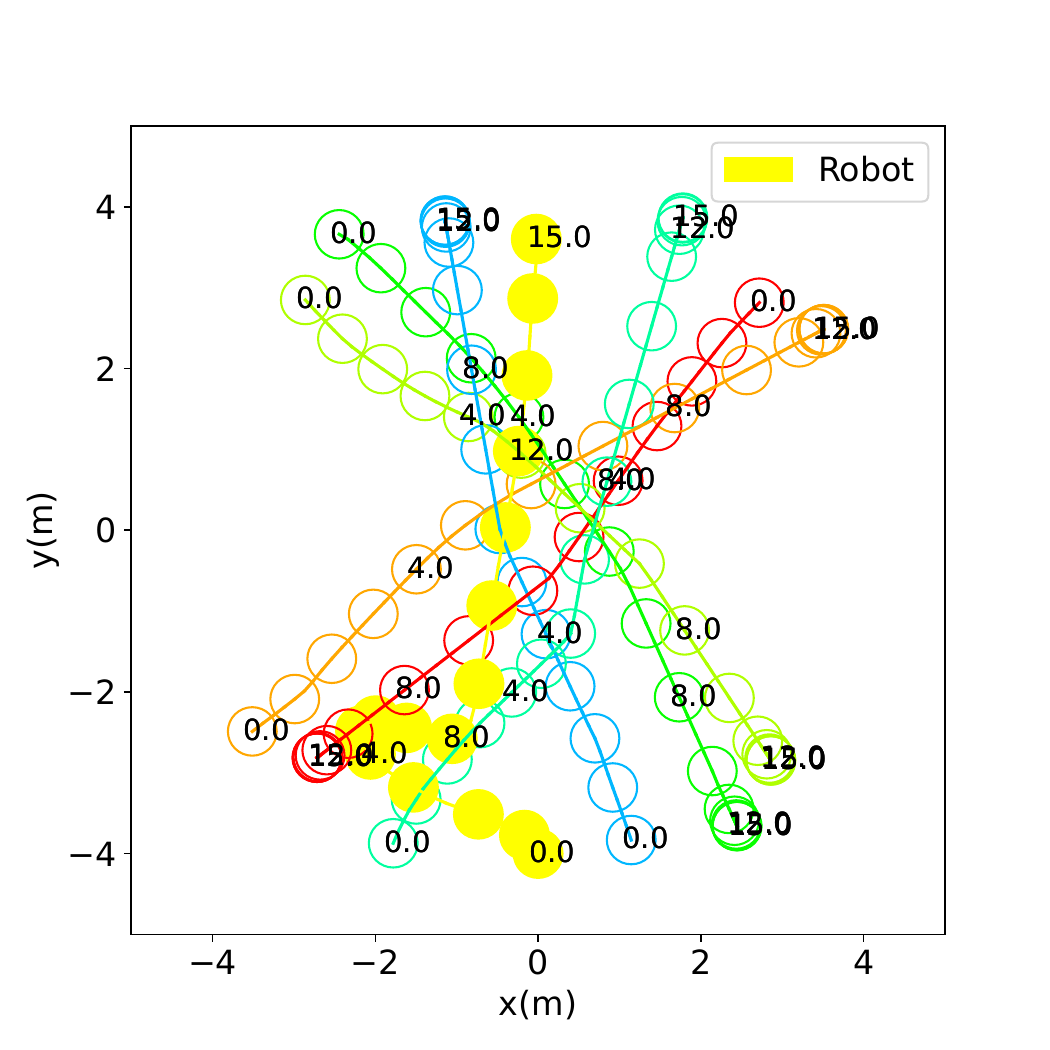}
        \caption{RPR-TD3BC}
        \label{RPR-TD3BC}
    \end{subfigure}
    \hfill
    \begin{subfigure}[b]{0.23\textwidth}
        \centering
        \includegraphics[width=\textwidth]{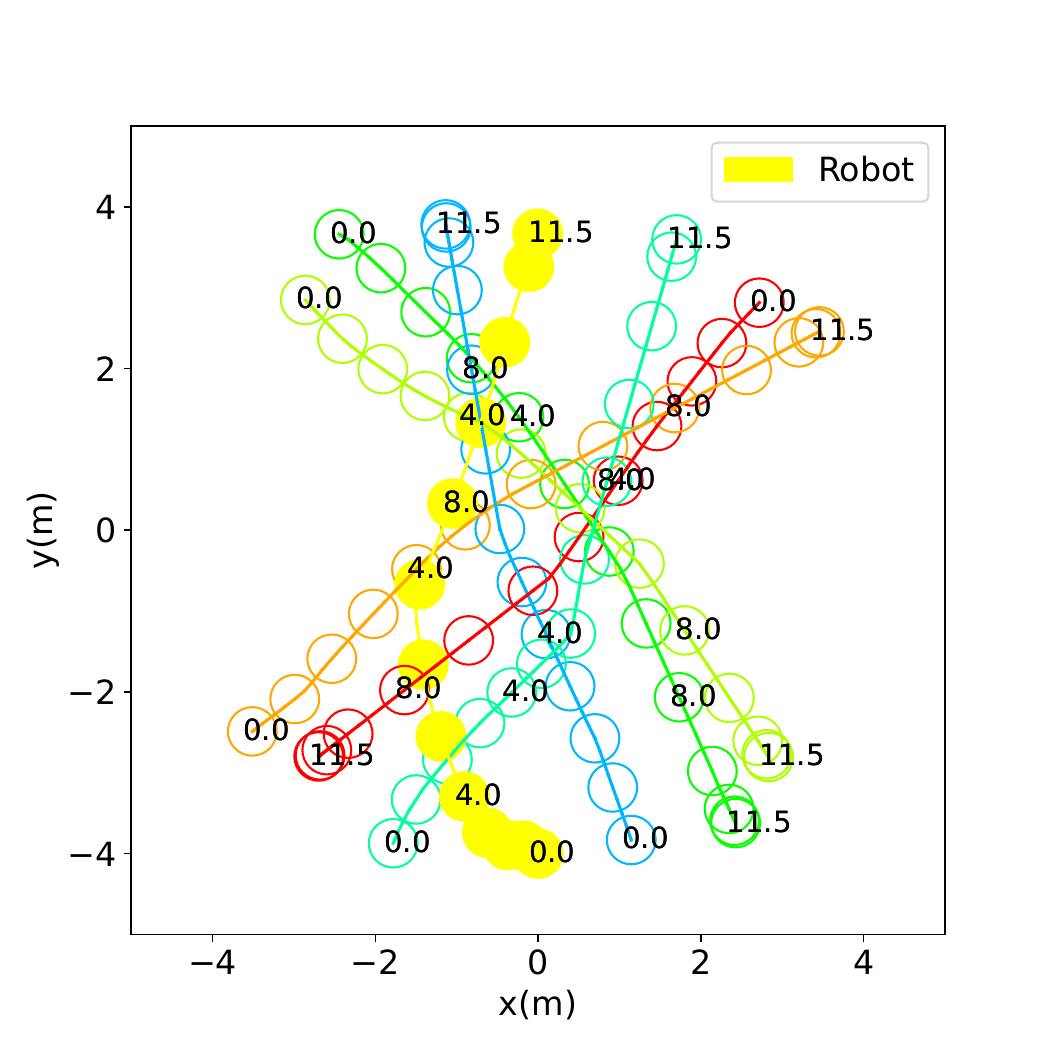}
        \caption{SPLC-TD3BC}
        \label{SPLC-TD3BC}
    \end{subfigure}

    \caption{Comparison of robot trajectories using different methods in identical crowd robot navigation test scenarios. Yellow indicates the robot's trajectory, while other colors represent pedestrian trajectories.}
    \label{fig4}
\end{figure*}

\subsection{Quantitative Evaluation}

This section details a comprehensive evaluation comparison through qualitative analysis, spotlighting vital metrics in the context of the crowd robot navigation.
These metrics involve "\textit{Success}", "\textit{Collision}", "\textit{Timeout}" and "\textit{Time}". In the qualitative evaluation, all
methods are evaluated in 500 testing cases. Their comparison results are listed in Table \ref{table3}.

\begin{table}[htbp]
\centering
\caption{Quantitative results of all methods}
\label{table3}
\begin{tabular}{lcccc}
\toprule
\textbf{Methods} & \textbf{Success} & \textbf{Collision} & \textbf{Timeout} & \textbf{Time} \\
\midrule
HR-IQL       & 93.40\% & 6.20\% &  0.40\%   & 11.95 s \\
HPR-IQL       & 93.20\% & 6.80\% &  0.00\%   & 11.10 s \\
RPR-IQL        & 92.20\% & 7.80\% &  0.00\%   & 11.06 s \\
SPLC-IQL       & 94.60\% & 5.40\% &  0.00\%   & 11.37 s \\
\midrule
HR-CQL       & 36.40\% & 63.60\% &  0.00\%   & 11.05 s \\
HPR-CQL       & 82.60\% & 17.40\% &  0.00\%   & 10.86 s \\
RPR-CQL        & 86.20\% & 13.80\% &  0.00\%   & 10.75 s \\
SPLC-CQL       & 95.40\% & 4.60\% &  0.00\%   & 11.18 s \\
\midrule
HR-TD3BC       & 82.40\% & 13.00\% &  4.60\%   & 13.70 s \\
HPR-TD3BC       & 62.20\% & 7.00\% &  30.80\%   & 16.74 s \\
RPR-TD3BC        & 79.00\% & 12.60\% &  8.40\%   & 14.63 s \\
SPLC-TD3BC       & 90.60\% & 7.40\% &  2.00\%   & 11.90 s \\
\bottomrule
\end{tabular}
\end{table}


As shown in the first four rows of Table \ref{table3}, HR-IQL yields the longest navigation time, as its manual reward function fails to capture those subtle and unspoken social norms inherent in the crowd robot navigation and instead focuses only on local sparse information, thereby leading to myopic behaviors. Although HPR-IQL can implicitly infer desirable navigation intentions from human judgments, subjectivity of human annotators incurs the reward bias, ultimately resulting in a lower success rate. RPR-IQL shows similar performance, providing slightly more stable navigation without significantly changing success rate. In contrast, our SPLC-IQL achieves consistent improvements across all evaluation metrics, indicating that the social preference feedback mechanism provides a more faithful quantification of broad social norms and effectively mitigates reward bias, thereby enhancing navigation success rate and efficiency.

In the fifth to eighth rows of Table \ref{table3},  HR-CQL exhibits a very low success rate due to the inherent conservativeness of the algorithm combined with biases in handcrafted rewards. HPR-CQL improves both success rate and navigation time, but it still produces suboptimal behaviors due to insufficient reward accuracy. RPR-CQL demonstrates further improvements, yielding higher success rates and shorter navigation times. Notably, our SPLC-CQL increases the success rate to 95.40\% while more effectively balancing efficiency and safety. The main reason is that the social preference feedback mechanism can effectively alleviate the reward bias, thereby making the CQL policy both more stable and effective.

In the last four rows of Table \ref{table3}, HR-TD3BC performs poorly in both success rate and navigation efficiency because handcrafted rewards fail to sufficiently amplify reward differences, keeping the policy in suboptimal behaviors. HPR-TD3BC performs even worse than handcrafted rewards, demonstrating that insufficient reward modeling cannot effectively guide the policy toward optimal behaviors. RPR-TD3BC improves over HPR-TD3BC by alleviating performance degradation, but remains inferior to handcrafted rewards. In contrast, our SPLC-TD3BC achieves the best performance across all metrics. This reveals that SPLC provides a more discriminative and reliable signal, effectively reducing reward bias and enabling the policy to overcome suboptimal behaviors and achieve superior navigation performance.

Overall, across the three offline RL methods, our SPLC algorithm better quantifies broad social norms and mitigates reward bias arising from the inherent unpredictability and often uncooperative dynamics of pedestrian motion.

\begin{figure*}[t]
    \centering
    \begin{subfigure}[t]{0.32\linewidth}
        \centering
        \includegraphics[width=\linewidth]{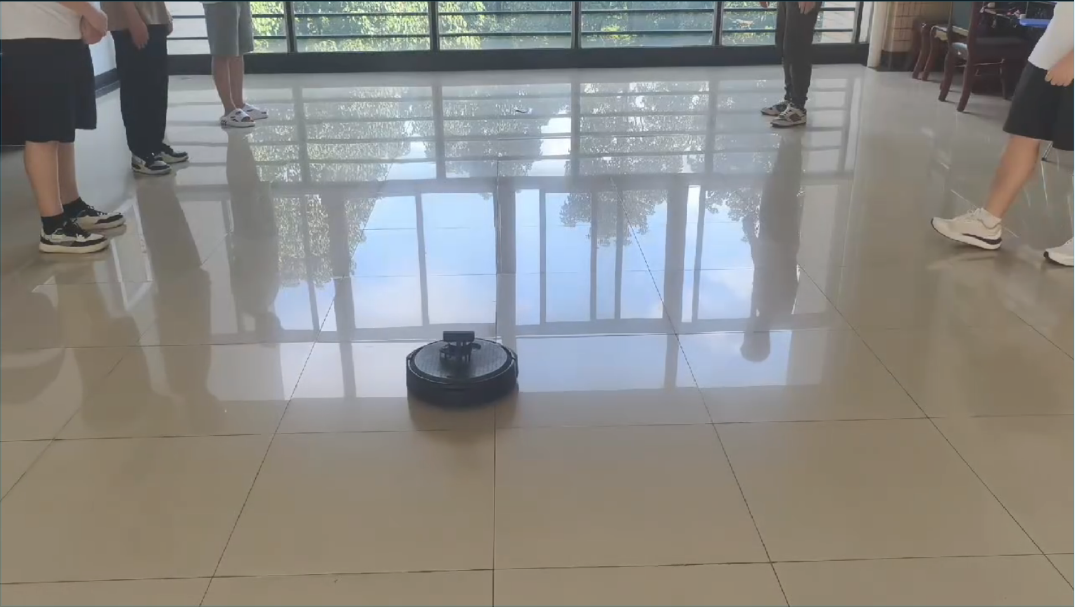}
        \caption{start time}
        \label{start}
    \end{subfigure}
    \hfill
    \begin{subfigure}[t]{0.32\linewidth}
        \centering
        \includegraphics[width=\linewidth]{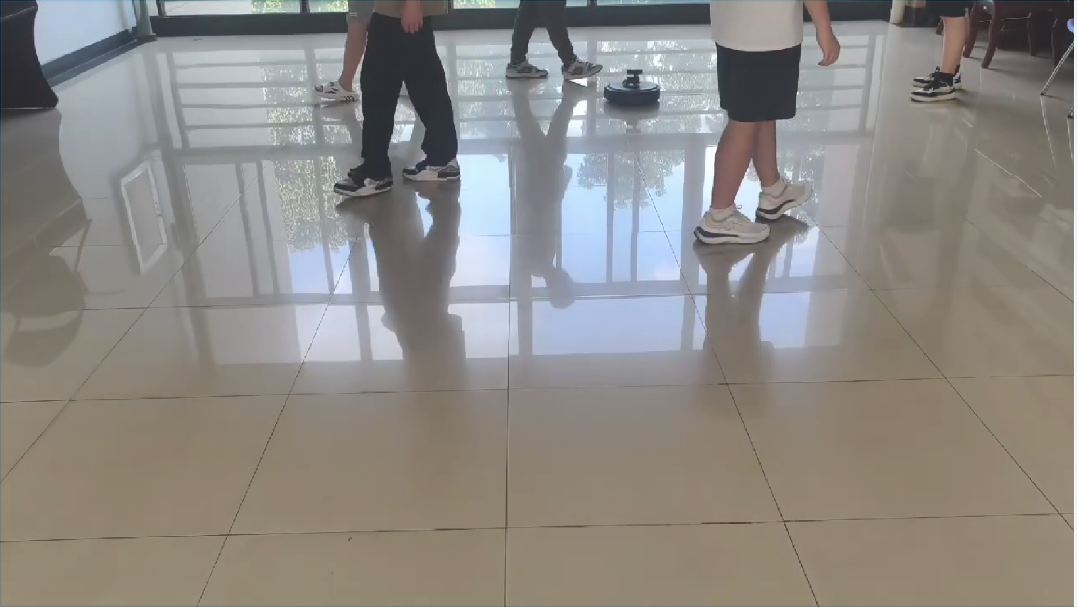}
        \caption{mid time}
        \label{mid}
    \end{subfigure}
    \hfill
    \begin{subfigure}[t]{0.32\linewidth}
        \centering
        \includegraphics[width=\linewidth]{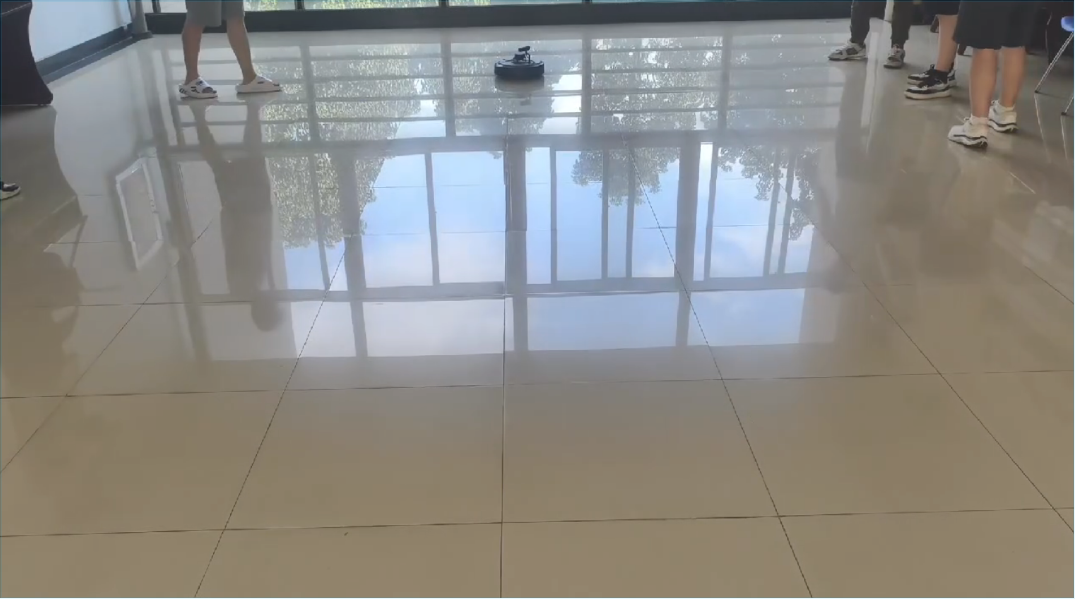}
        \caption{end time}
        \label{end}
    \end{subfigure}
    \caption{The three sub-figures sequentially show the robot’s navigation process in the real-world experiment of our algorithm.}
    \label{real}
\end{figure*}

\subsection{Qualitative Evaluation}
To qualitatively validate our algorithm, global trajectories from different methods are compared to visualize robot behaviors under a shared initial condition and target formation. Details are shown in Fig.~\ref{fig4}. 

As illustrated in Figs.~\ref{IQL}-\ref{RPR-IQL}, HR-IQL, HPR-IQL and RPR-IQL execute rightward turns at 4 seconds to avoid pedestrians, but subsequently enter densely crowded areas, resulting in slow subsequent movement. This reflects myopic and unnatural navigation caused by inaccurate rewards. In contrast, Fig.~\ref{SPLC-IQL} shows that our SPLC-IQL proactively adjusts the path by selecting the opposite side, thereby steering the robots toward a sparser and safer region and completing the navigation task more efficiently, which highlights that our SPLC better quantifies social norms and incorporates global awareness into the navigation policy, enabling the agent to anticipate the collective flow of the crowd and maintain appropriate social distancing while avoiding disruptive interactions.

Next, in Fig.~\ref{CQL}, the trajectory of HR-CQL collides at 4.2 seconds, as the robot trained with handcrafted rewards lacks sufficient obstacle avoidance capability. Fig.~\ref{HPR-CQL} and~\ref{RPR-CQL} show that HPR-CQL and RPR-CQL avoid direct collision but still approach pedestrians too closely at 4 seconds, with a tendency to pass through rather than detour. In Fig.~\ref{SPLC-CQL}, our SPLC-CQL instead performs a clear rightward bypass, ensuring safer navigation and more stable trajectory evolution. These comparisons indicate that our preference evaluation criteria fully account for the intricacies of pedestrian dynamics, providing CQL with more stable and task-aligned learning signals, and thereby enhancing both the safety and efficiency of navigation with more socially compliant yielding behavior.

Furthermore, Fig.~\ref{TD3BC} shows that HR-TD3BC completes the task but slows down significantly in dense crowds, exhibiting inefficient low-speed detours. Similarly, Fig.~\ref{HPR-TD3BC} and~\ref{RPR-TD3BC} indicate that HPR-TD3BC and RPR-TD3BC are overly conservative, prolonging navigation time and sacrificing efficiency. By comparison, Fig.~\ref{SPLC-TD3BC} demonstrates that our SPLC-TD3BC anticipates dense pedestrian regions and executes early avoidance, maintaining a safe distance from pedestrians during motion, thereby completing the navigation task more safely and in a shorter duration. The main reason is that our principled preference evaluation criteria better quantify social norms in crowd robot navigation and mitigate reward bias arising from the inherent unpredictability and often uncooperative dynamics of pedestrian motion, resulting in more socially compliant navigation.

\subsection{Real-world Experiments}
This work was further validated through real-world experiments in addition to simulations. The experimental platform comprised a laptop with an R9-7940HX processor and an RTX 4060 GPU, integrated with a TurtleBot4. An RPLIDAR-A1 LiDAR combined with a pedestrian leg detection algorithm was employed to identify pedestrians and estimate their relative positions and velocities. The robot’s state was tracked using its built-in chassis odometry after mapping the environment. During the experiments, the collected state data were transmitted to the laptop, where the deployed SPLC-IQL algorithm generated the control actions, which were then executed by the robot in the physical environment.

As illustrated in Fig.~\ref{real}, the robot successfully reached the target while avoiding collisions with five pedestrians. In Fig.~\ref{mid}, it can be observed that the robot exhibited clear obstacle-avoidance behaviors when pedestrians approached nearby. Full real-world demonstrations of the proposed algorithm are provided in the submitted multimedia materials. These results confirm that the algorithm can be effectively transferred from simulation to a physical robot, enabling it to accomplish crowd robot navigation tasks.

\section{CONCLUSIONS}

In this study, we have proposed the SPLC algorithm to address the reward function design issue in DRL-based crowd navigation. Our SPLC introduces a social preference feedback mechanism that automatically generates preference data via preference evaluation criteria to model the reward function, alleviating reward biases in the crowd navigation. By integrating SPLC with several offline RL algorithms, we demonstrated through extensive experiments that the proposed SPLC algorithm achieves superior performance in terms of both average navigation success rate and navigation efficiency. Finally, experiments on the TurtleBot4 show that our SPLC algorithm successfully transfers from simulation to real-world robots. In future work, we will conduct more systematic real-world experiments to further examine sim-to-real transfer, while incorporating real-world crowd navigation datasets to improve the robustness and practical applicability of the proposed SPLC framework.

\bibliographystyle{IEEEtran}
\bibliography{bibtex}

\end{document}